\documentclass[10.5pt,compsoc]{BD}
\usepackage{graphicx}
\usepackage{footmisc}
\usepackage{subfigure}
\usepackage{url}
\usepackage{multirow}
\usepackage[noadjust]{cite}
\usepackage{amsmath,amsthm}
\usepackage{amssymb,amsfonts}
\usepackage{booktabs}
\usepackage{color}
\usepackage{ccaption}
\usepackage{booktabs}
\usepackage{float}
\usepackage{fancyhdr}
\usepackage{caption}
\usepackage{xcolor,stfloats}
\usepackage{comment}
\graphicspath{{figures/}}
\usepackage{cuted}  
\usepackage{captionhack}
\usepackage{epstopdf}
\UseRawInputEncoding

\headevenname{\normalsize{\textbf{\emph{Big Data Mining and Analytics,} xxxxxxx}} 24xx, x(x): xxx-xxx}%
\headoddname{{\sf Ma et al.:}\quad {\textbf{\emph{Autism Spectrum Disorder Classification...}}}}%

\newtheoremstyle{mystyle}{0pt}{0pt}{\normalfont}{1em}{\bf}{}{1em}{}
\theoremstyle{mystyle}
\renewcommand\figurename{Fig.~}

\newcommand{\nop}[1]{}
\usepackage{amsmath}

\makeatletter
\renewcommand{\@biblabel}[1]{[#1]\hfill}
\makeatother

\begin{document}

\begin{strip}
	{\center
		{\LARGE\textbf{Autism Spectrum Disorder Classification with Interpretability in Children based on Structural MRI Features Extracted using Contrastive Variational Autoencoder}
			\vskip 9mm}}
	
	{\center {\sf \large
			Ruimin Ma $^\dagger$, Ruitao Xie $^\dagger$, Yanlin Wang, Jintao Meng, Yanjie Wei, Yunpeng Cai, Wenhui Xi$^*$, and Yi Pan$^*$
		}
		\vskip 5mm}

	\centering{
		\begin{tabular}{p{160mm}}
			{\normalsize
				\linespread{1.6667} %
				\noindent
				\bf{Abstract:} {\sf
					Autism spectrum disorder (ASD) is a highly disabling mental disease that brings significant impairments of social interaction ability to the patients, making early screening and intervention of ASD critical. With the development of the machine learning and neuroimaging technology, extensive research has been conducted on machine classification of ASD based on structural Magnetic Resonance Imaging (s-MRI). However, most studies involve with datasets where participants' age are above 5 and lack interpretability. In this paper, we propose a machine learning method for ASD classification in children with age range from 0.92 to 4.83 years, based on s-MRI features extracted using contrastive variational autoencoder (CVAE). 78 s-MRIs, collected from Shenzhen Children's Hospital, are used for training CVAE, which consists of both ASD-specific feature channel and common shared feature channel. The ASD participants represented by ASD-specific features can be easily discriminated from TC participants represented by the common shared features. In case of degraded predictive accuracy when data size is extremely small, a transfer learning strategy is proposed here as a potential solution.  Finally, we conduct neuroanatomical interpretation based on the correlation between s-MRI features extracted from CVAE and surface area of different cortical regions, which discloses potential biomarkers that could help target treatments of ASD in the future.}
				}
			
		\end{tabular}
	}
\vskip 6mm

\vskip -3mm
\small\end{strip}

\thispagestyle{plain}%
\thispagestyle{empty}%
\makeatother
\pagestyle{tstheadings}

\begin{figure*}[b]
	\vskip -6mm
	\begin{tabular}{p{108mm}}
		\toprule\\
	\end{tabular}
	\vskip -4.5mm
	\noindent
	\setlength{\tabcolsep}{1pt}
	\begin{tabular}{p{170mm}}
		
		$\bullet$ Ruimin Ma, Yanlin Wang, Jintao Meng, Yanjie Wei, Yunpeng Cai, Wenhui Xi are with Shenzhen Institute of Advanced Technology, Chinese Academy of Sciences, Shenzhen 518055, China. Email: ruiminma09@gmail.com; simon971234@gmail.com; jt.meng@siat.ac.cn; yj.wei@siat.ac.cn; yp.cai@siat.ac.cn; wh.xi@siat.ac.cn.\\
		$\bullet$ Ruitao Xie is with Shenzhen Institute of Advanced Technology, Chinese Academy of Sciences, Shenzhen 518055, China, and also with University of Chinese Academy of Sciences, Beijing 100049, China. Email: rt.xie@siat.ac.cn.\\
		$\bullet$ Yi Pan is with Shenzhen Key Laboratory of Intelligent Bioinformatics, and also with Center for High Performance Computing, Shenzhen Institute of Advanced Technology, Chinese Academy of Sciences, Shenzhen 518055, China. Email: yi.pan@siat.ac.cn.\\
		$\dagger$ Ruitao Xie and Ruimin Ma contribute equally to this paper.\\
		$*$ To whom correspondence should be addressed.\\
		

	\end{tabular}
\end{figure*}\large

\vspace{3.5mm}
\section{Introduction}
\label{s:introduction}
\begin{figure}
	\centering
	\includegraphics[scale=1.3]{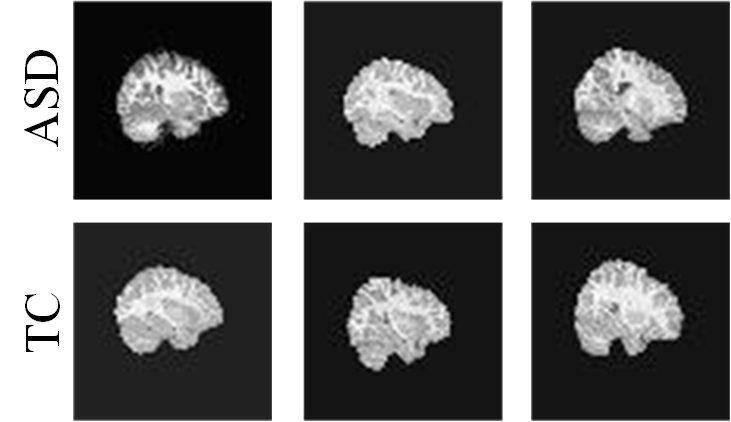}
	\caption{Several ASD and typical control (TC) examples used in our work.}
	\label{cases}
\end{figure}

\noindent
Autism spectrum disorder (ASD) is a highly disabling mental disease, the patients of which have impaired social interaction ability, verbal and nonverbal communication deficiencies or other problems$^{[1]}$. The diagnosis of autism is usually based on behavior scores evaluated by doctors, which may result in inaccurate diagnosis of ASD$^{[2]}$. Machine learning technology has aroused great attention recently and its role in analyzing medical data has been proven to be effective and efficient in many tasks of computer-aided medical diagnosis, including ASD detection. Accurate machine classification of ASD based on structural magnetic resonance imaging (s-MRI, a medical imaging technique that uses magnetic fields and radio waves to generate detailed images of the internal structures of the body, particularly the brain) becomes popular and critical recently$^{[3]}$. In this paper, we also perform ASD classification study using brain s-MRI, and here we show several examples used in our work in Figure \ref{cases}.

Most researchers conduct machine classification of ASD based on s-MRI collected from the Autism Brain Imaging Data Exchange (ABIDE)$^{[4, 5]}$. Kong et al. construct an individual brain network to extract connectivity features between each pair of regions of interest from the s-MRI and select top 3000 connectivity features for ASD classification via a deep neural network classifier$^{[6]}$. Hiremath et al. design a dense model with 4 dense blocks to extract deep features of the s-MRI, which are then combined with the morphometric hand-crafted features, i.e., volume and surface features of different brain parcellations, and fed into a gender specific linear discriminant classifier, to classify ASD from typical control (TC)$^{[7]}$. Gao et al. construct individual-level morphological covariance brain network according to SRI24 atlas$^{[8]}$ and gray matter volume map of each subject's s-MRI, which is used as the input for the ResNet-based classifier for ASD classification$^{[9]}$. Yang et al.$^{[10]}$ use Class Activation Mapping (CAM)$^{[11]}$ and Gradient-weighted CAM (GradCAM)$^{[12]}$ to design three models with s-MRIs as inputs for classifying ASD. Mishra et al. average the outputs produced by two models, which have the same deep convolution network architecture but are trained using different optimizers (Adam and Nadam), for ASD classification based on s-MRI inputs$^{[13]}$. Sharif et al. employ a pre-trained VGG16 model for ASD classification using features extracted from corpus callosum and intracranial brain volume of s-MRI$^{[14]}$. Wang et al. propose a transformer-based framework for ASD classification, which makes the extraction of both local and global information from the s-MRI possible$^{[15]}$. Obviously, ABIDE serves as a valuable benchmark for machine classification of ASD based on s-MRI, however, the samples in ABIDE are all above the age of five, which makes it difficult to conduct early (such as under the age of five) screening research for ASD. Besides, most of these studies has mediocre classification accuracies, which may potentially be attributed to the heterogeneous data collected from different resources$^{[16]}$.

Few studies use s-MRI from other datasets for machine classification of ASD. Sarovic et al. develop a user-friendly multivariate statistical method for analyzing the s-MRI of  24 ASD participants (mean age $\pm$ standard deviation: 30.6 $\pm$ 7.1 years) and 21 TC participants (28.2 $\pm$ 6.4 years$^{[17]}$).  Conti et al. extract features from s-MRI of 18 TC children (55 $\pm$ 13 months) and 26 ASD children (56 $\pm$ 11 months), and then perform multivariate analysis based on support vector machine binary classifiers$^{[18]}$. Zhang et al.$^{[19]}$ propose a Siamese verification model for ASD classification, where 30 ASD and 30 TC are randomly selected from National Database of Autism Research (NDAR)$^{[20]}$ for the experiment with their longitudinal brain s-MRI at 6 and 12 months of age. Early-stage machine classification of ASD are conducted in some of those studies mentioned above; however, all these works have mediocre performance, with predictive accuracies 73.2\% for Sarovic et al.$^{[17]}$, 87\% for Zhang et al. $^{[19]}$ and 0.73 Area Under the Curve (AUC) for Conti et al.$^{[18]}$. In addition, most existing machine learning methods for ASD classification lack interpretability study, which hinders their clinical application.

In this work, we adopt contrastive variational autoencoder (CVAE) to extract the ASD-specific features for ASD detection using 78 s-MRIs (42 ASD and 36 TC) collected from Shenzhen Children's Hospital. The age of the children involved in the study ranges between 0.92-4.83-year-old. The ASD-specific features are extracted from the CVAE after the training for ASD classification, the mean accuracies of which are constantly above 0.94 in different cross-validation scenarios. A transfer learning strategy is introduced as well for the scenario when high accuracy won't be achieved with extremely limited amount of data. Besides the ASD classification, neuroanatomical interpretation is thoroughly conducted, to identify potential biomarkers that could help target clinical treatment for ASD. This work can serve as a good benchmark for ASD study in children based on s-MRI.

\section{Methods}
\label{s:Methods}
\noindent
\subsection{Overall framework}
\begin{figure*}
	\centering
	\includegraphics[scale=0.52]{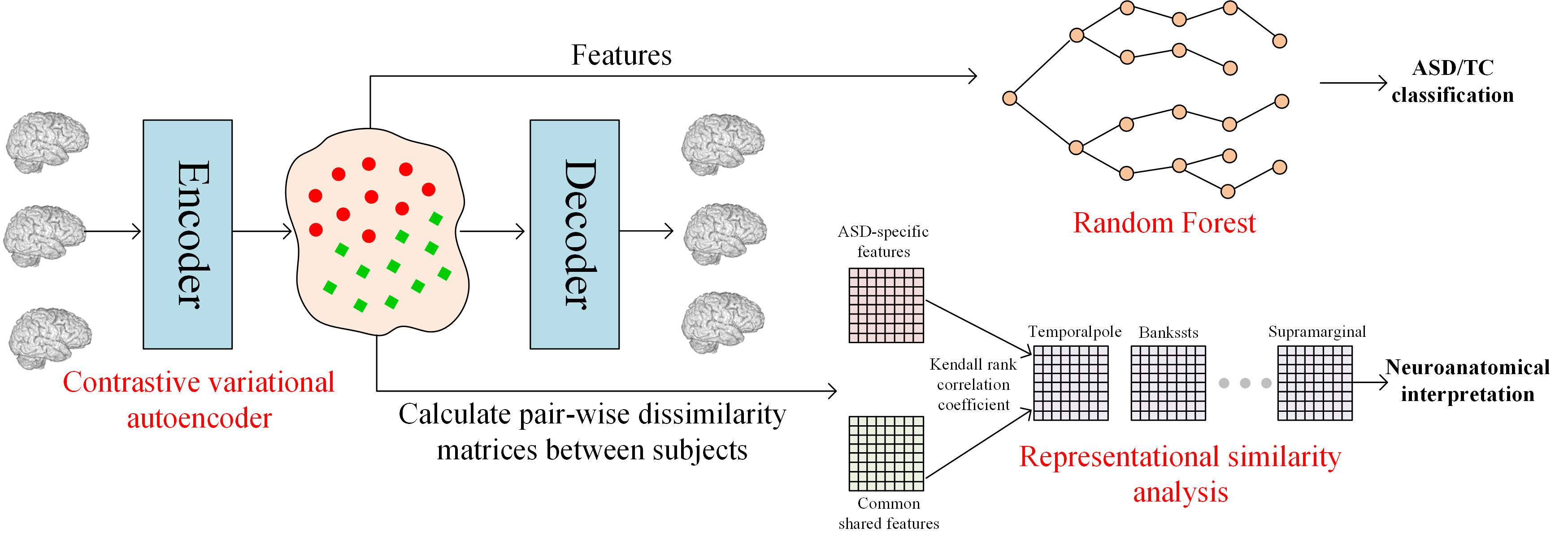}
	\caption{Overall framework of our proposed algorithm for ASD classification and interpretability study.}
	\label{overall-framework}
\end{figure*}

In this paper, we propose an algorithm that simultaneously achieves ASD classification and neuroanatomical interpretability based on contrastive variational autoencoder (CVAE). Specifically, we use CVAE to separate ASD-specific and common shared information in brain imaging, obtaining more targeted features. Based on the learned feature space, we employ the random forest algorithm for ASD classification. In addition, we compute the relationships between each pair of subjects through ASD-specific and common shared features respectively, and obtain two pair-wise dissimilarity matrices. By comparing these matrices with the matrices representing relationships between each pair of subjects with respect to different brain regions, we can identify brain regions with high correlation to ASD-specific features, achieving interpretability and facilitating the exploration of ASD biomarkers. The specific flowchart is shown in Figure \ref{overall-framework}.

\subsection{Dataset}
\noindent
In this study, we obtained 78 structural MRIs (s-MRIs) from Shenzhen Children's Hospital, where 42 of them are obtained from ASD participants (10 female, 32 male) and 36 of them are obtained from TC participants (19 female, 17 male). Age (in years) ranges from 1.00 to 4.83 in ASD group and from 0.92 to 4.33 in TC group. The s-MRI used as inputs for the contrastive machine learning remain in native space, i.e., they are not post-processed using any specific brain templates and are normalized between 0 and 1 and then resampled to a resolution of 64 $\times$ 64 $\times$ 64 before being fed into the CVAE for feature extraction. While in Section 3.3, the s-MRI are post-processed into features of different cortical regions: for children between 0-2-year-old, the children's s-MRIs are processed by the Infantfreesurfer$^{[21]}$ tools, based on the per-month strategy (i.e., the 0-2-year-old, range from 0 to 24 months, are divided into 24 intervals, with each interval representing the length of a month, and children's s-MRIs within each interval are processed separately); for children between 2-5-year-old, the children's s-MRIs are first projected onto the age-specific brain templates to create the prior masks using ANTs$^{[22]}$, and then are further processed by Freesurfer$^{[23]}$ tool; the Freesurfer tools automatically parcellate the cortex and assign a neuroanatomical label to each location of the cortex (i.e., gray matter) based on probabilistic information, and the cortex is divided into 34 regions in this study based on Desikan-Killiany atlas$^{[24]}$; surface area of each cortical region is measured as the feature indicating its uniqueness. All data used in this study are approved by local Internal Review Boards and carried out in accordance with ethics guidelines and regulations, and informed consents are obtained from all participants, or parent/legal guardians if participants are under 18.

\subsection{Contrastive variational autoencoder (CVAE)}
\begin{figure}
	\centering
	\includegraphics[scale=0.56]{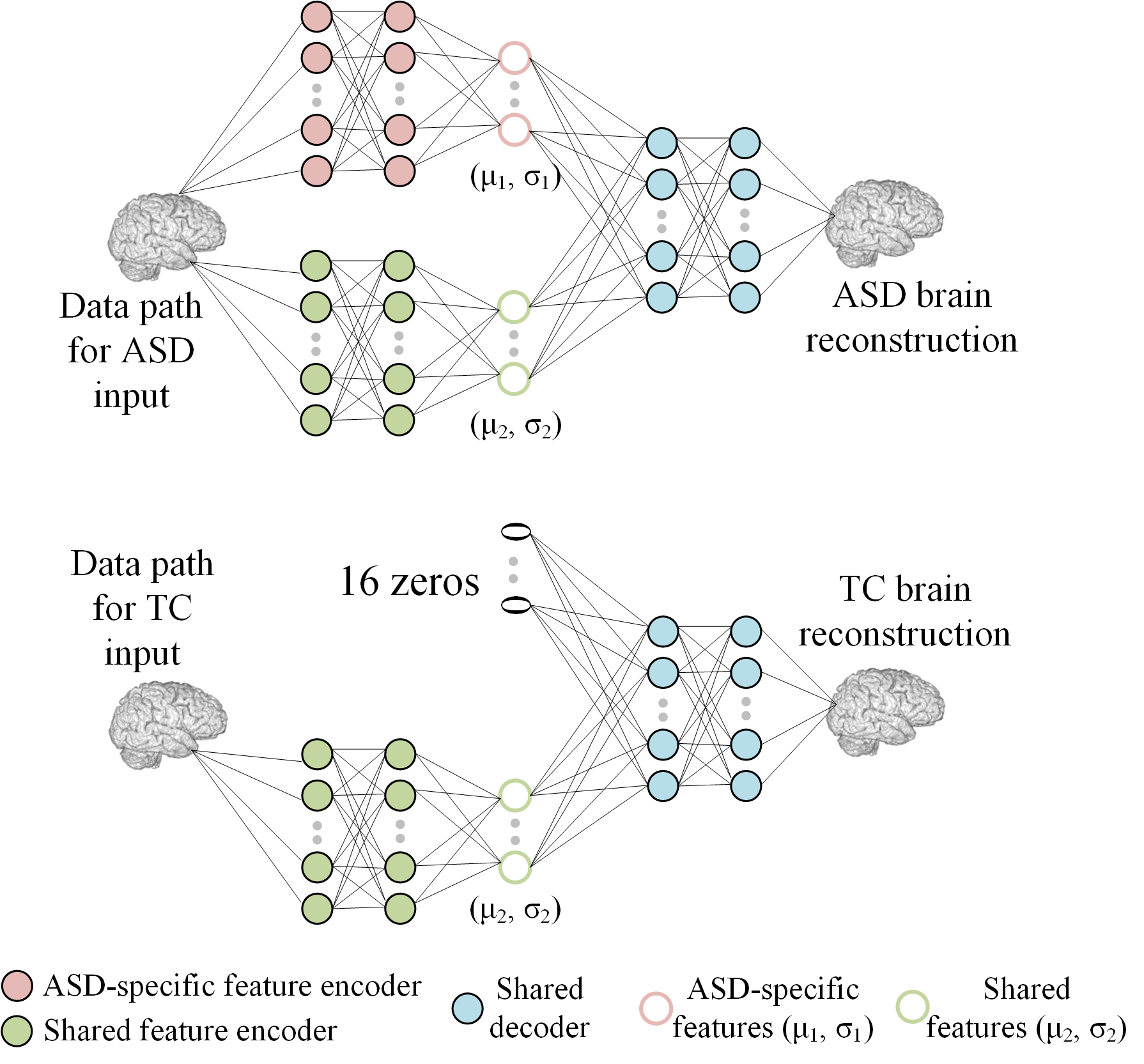}
	\caption{Contrastive variational autoencoder for extracting ASD-specific features and common shared features.}
	\label{CVAE}
\end{figure}

\noindent
The CVAE$^{[25, 26]}$ is used in our work to extract ASD-specific features as well as common shared features. It is a probabilistic-latent-variable model applied to contrastive analysis, which uses specific and shared feature extraction networks to emphasize the latent features of interest while suppressing uninteresting features in the data. As shown in Figure \ref{CVAE}, the CVAE model consists of two encoders with identical architectures to extract ASD-specific and common shared features separately, and one shared decoder to reconstruct s-MRI of participants' brains. The encoder each has two convolutional layers (kernel size: 3, stride size: 2, number of filters: 64 and 128 respectively), and two modules each consisting of two fully connected layers. In these two modules, the first layers are shared and include 128 output nodes and Relu activation operation, while the second layers (both with 16 output nodes) are independent and designed for the final extraction of the mean values (denoted as $\mu$) and variance values (denoted as $\sigma$) of the features separately. The decoder has two fully connected layers both with Relu operation (128 and 524288 output nodes respectively) and three deconvolutional layers (kernel size: 3, stride size: 2, number of filters: 32, 16 and 1 respectively). The s-MRIs of TC participants pass through the shared feature encoder only, while the s-MRIs of ASD participants pass through both ASD-specific and shared feature encoders, which learn the distribution (mean and variance) of the features. The dimensions of both ASD-specific features and common shared features are 16. When reconstructing ASD brains, both common shared features and ASD-specific features (sampled from the features distribution learned from the encoders) are fed into the decoder, while reconstructing TC brains, only common shared features are sampled and then concatenated with 16 zeros as inputs for the same decoder. 

\subsection{Implementation details}
\noindent
All 78 s-MRIs (reshaped as 64 $\times$ 64 $\times$ 64 pixel) from Shenzhen Children's Hospital are leveraged for training CVAE, with Adam$^{[27]}$ adopted, where the parameters of the learning rate ($lr$, controls the step size of parameter updates during network training), $\beta_{1}$ (the exponential decay rate for the first moment estimates), $\beta_{2}$ (the exponential decay rate for the second moment estimates) and $\epsilon$ (a small constant that is added to the denominator of the gradient estimates to prevent division by zero) are set as $10^{-3}$, 9 $\times$ $10^{-1}$, 9.99 $\times$ $10^{-1}$ and $10^{-7}$ respectively. The training batch size is 8, which means that for each iteration of training, 8 s-MRIs of ASD participants and 8 s-MRIs of TC participants, randomly sampled from the population, are fed into the network for learning. The training ends when the mean-square-error between input s-MRIs and reconstructed s-MRIs falls below 5 $\times$ $10^{-4}$ for Section 3.1, and 5 $\times$ $10^{-3}$ for Section 3.2. The experiments are conducted using Python 3.7 environment with the TensorFlow 1.13 framework and accelerated by 2 GPU cards. 

\section{RESULTS}
\label{s:RESULTS}
\noindent
\subsection{ASD-specific features for ASD classification}
\begin{figure*}
	\centering
	\includegraphics[scale=0.42]{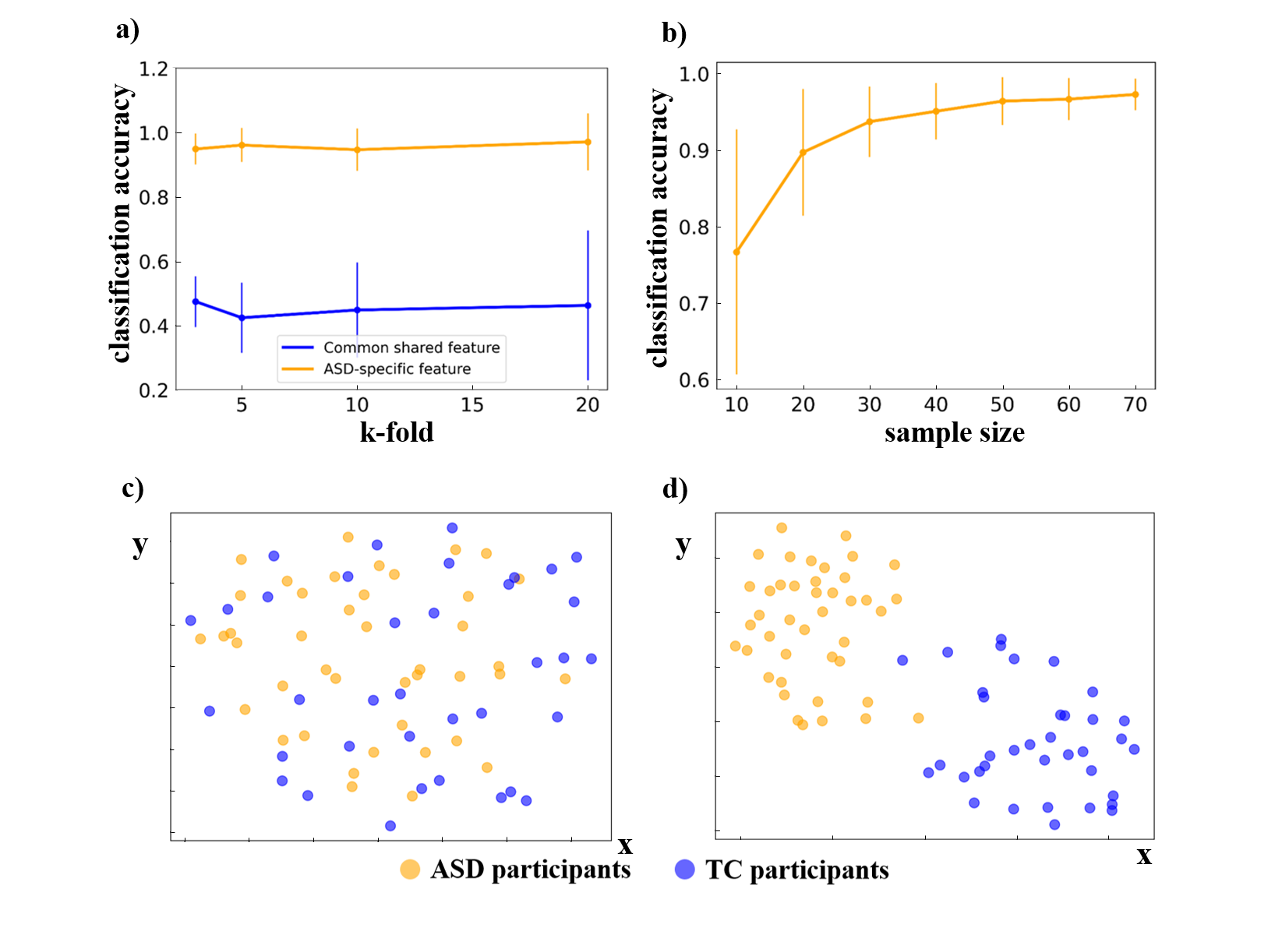}
	\caption{Classification results and the visualization  analysis of the features. a) accuracy of classifying ASD with ASD participants being represented either by ASD-specific features or common shared features; b) relationship between classification accuracy and sample size, the accuracy of classifying ASD was measured based on ASD participants being represented by ASD-specific features; visualizing ASD participants and TC participants in 2-D feature space with TC participants being represented by common share features while ASD participants being represented by either c) common shared features, or d) ASD-specific features.}
	\label{classification1}
\end{figure*}

\noindent
According to the methods mentioned in Section 2.3, the ASD participant can be represented by the ASD-specific features and common shared features, where the ASD-specific features reveal the irregular brain developments that never happen in the TC participants, thus serving the purpose of classifying ASD. We first evaluate the quality of ASD-specific features in ASD classification, and the results are shown in Figure \ref{classification1}. The ASD participants are either represented by ASD-specific features or common shared features, while the TC participants are represented by the common shared features. A random forest$^{[28]}$ model (with number of trees being set to 100) is introduced to classify ASD based on the features defined above, and the classification accuracies are measured in K-fold cross-validations. We try different K-fold numbers to create varying train/test scenarios, so that the comparison between ASD-specific features and common shared features is more reliable. According to Figure \ref{classification1}.a, the classification accuracy is much better improved when ASD-participants are represented using ASD-specific features, however the K-fold number changes (numerical accuracy results were presented in Table \ref{accuracy}).  Besides, it's also understandable that increment of K-fold number leads to increment of variance in classification accuracy (characterized by the error bar in Figure \ref{classification1}.a), since there are more test scenarios when K-fold number is larger. 

\begin{table*}[tt]
	\caption{Mean accuracies ($\pm$std) of classifying ASD with ASD participants being represented by ASD-specific features or common shared features with different K-fold numbers.}
	\vskip 5mm
	\label{accuracy}
	\centering
		\begin{tabular}{lcccc}
			\toprule
			\multirow{2}*{Features used for classification}  &\multicolumn{4}{c}{K-fold cross-validations}  \\
			\cline{2-5}
			~    &3-fold &5-fold &10-fold &20-fold \\
			\midrule
			\emph{Common shared features}   & 0.4744$\pm$0.0790 & 0.4242$\pm$0.1094 & 0.4482$\pm$0.1476 & 0.4625$\pm$0.2334 \\
			\emph{ASD-specific feature}  &0.9487$\pm$0.0480 &0.9608$\pm$0.0529 &0.9464$\pm$0.0659 &0.9708$\pm$0.0885\\
			
			\bottomrule
		\end{tabular}
\end{table*}

In Figure \ref{classification1}.b, we evaluated the classification accuracies' dependence on sample size. For a specific sample size, e.g., 10, we sample 10 s-MRIs out of the total 78 and then we train a classification random forest on it to get the accuracy, where the ASD participants are represented using ASD-specific features. We repeat it for 100 times to get the error bar of classification accuracy (measured via standard deviation), to capture the uncertainty of classification accuracy's dependence on sample size.  The classification accuracy increases as sample size increases, which is intuitive since more data brings in more information. However, according to another study that classifies ASD via s-MRI features$^{[16]}$, negative relationship is observed between sample size and classification accuracy, which is attributed to the varying data quality caused by heterogeneous data resources. Relatively speaking, the results of Figure \ref{classification1}.b demonstrates a uniform data-quality distribution of data used here. 

To visually characterize the superiority of ASD-specific features in ASD classification, we visualize the common shared features of ASD participants in Figure \ref{classification1}.c and the ASD-specific features of ASD participants in Figure \ref{classification1}.d, with common shared features of TC participants as the backgrounds in both plots. The ASD-specific features and common shared features of ASD participants and common shared features of TC participants are both in high dimension originally, we use t-SNE$^{[29]}$ visualization technique to project them onto 2-D space. From Figure \ref{classification1}.c and \ref{classification1}.d, the ASD participants represented by ASD-specific features can be clearly separated from TC participants, while ASD participants represented by common shared features overlap well with TC participants. 

In order to validate the powerful feature extraction capability of our CVAE model, we conducted ablation study. We used the same random forest classification algorithm, but different inputs were used for training and testing the random forest model. Specifically, we used original images as inputs to the model. The experimental results are shown in Table \ref{Ablation}. The experimental results indicate that the features extracted by our proposed CVAE algorithm are more advantageous for ASD classification.

\begin{table*}[tt]
	\caption{Mean accuracies ($\pm$std) of classifying ASD using original images and features extracted from CVAE as input.}
	\vskip 5mm
	\label{Ablation}
	\centering
		\begin{tabular}{lcccc}
			\toprule
			\multirow{2}*{Different inputs for classification}  &\multicolumn{4}{c}{K-fold cross-validations}  \\
			\cline{2-5}
			~    &3-fold &5-fold &10-fold &20-fold \\
			\midrule
			\emph{Original images}   & 0.6282$\pm$0.1307 & 0.6675$\pm$0.1051 & 0.6679$\pm$0.1781 & 0.6667$\pm$0.2622 \\
			\emph{Features using CVAE (ours)}  &0.9487$\pm$0.0480 &0.9608$\pm$0.0529 &0.9464$\pm$0.0659 &0.9708$\pm$0.0885\\
			
			\bottomrule
		\end{tabular}
\end{table*}

To validate the superiority of our proposed algorithm, we conducted comparative experiments with several recent related works using ten-fold cross-validation. The experimental results are shown in Table \ref{compare}, which indicate that our proposed algorithm performs better in ASD classification.

\begin{table}
	\caption{Mean accuracies ($\pm$std) of classifying ASD using different methods.}
	\vskip 5mm
	\label{compare}
	\centering
	\begin{tabular}{lc}
		\toprule
		Methods   &Accuracy\\
		\midrule
		\emph{Jönemo et al.$^{[30]}$}  & 0.7571$\pm$0.1619 \\
		\emph{Sharif et al.$^{[14]}$} & 0.8214$\pm$0.0867 \\
		\emph{Nogay et al.$^{[31]}$}  &0.7214$\pm$0.1422 \\
		\emph{CVAE (ours)}   &0.9464$\pm$0.0659 \\
		\bottomrule
	\end{tabular}
\end{table}

\subsection{Transfer learning study of ABIDE-I}
\noindent
Collecting high-quality s-MRI data for ASD study in children is both technically and ethically difficult, especially for children under age six$^{[32, 33]}$ thus leading to a small dataset for exploration. As can be seen from the Figure \ref{classification1}.b, the average classification accuracy can't reach 0.90 when the size of data was below 30, causing a less predictive model. Classification performance of one specific task with limited training data could be generally improved by transferring the knowledge learned from a similar task with adequate data to this specific task, often noted as transfer learning$^{[34, 35]}$. ABIDE-I$^{[4]}$ is an open-sourced database containing well-recorded s-MRI for both ASD and TC participants, the knowledge of which can be potentially transferred to improve the classification accuracy of ASD detection in our work. 

\begin{figure}
	\centering
	\includegraphics[scale=0.3]{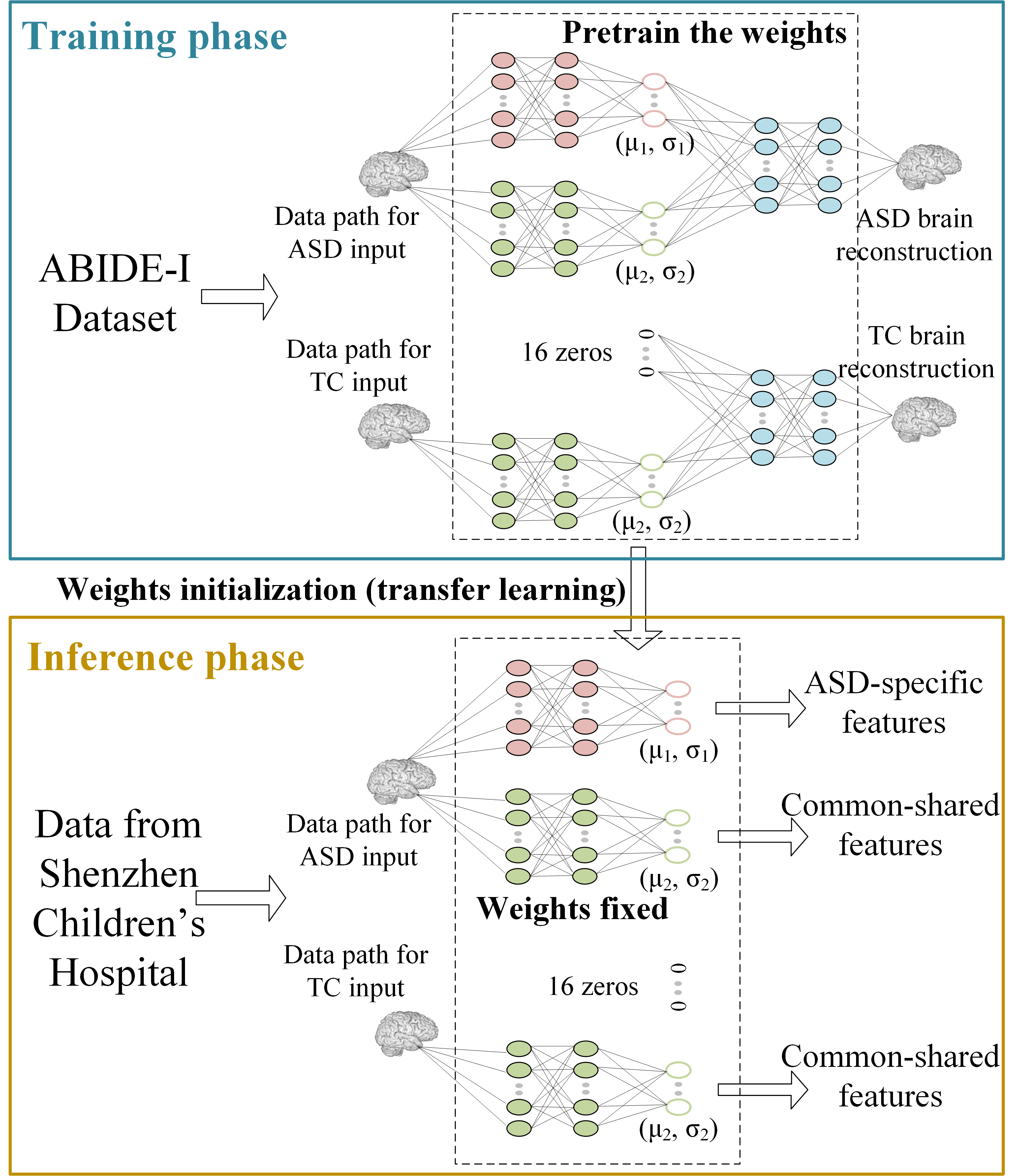}
	\caption{The schematic diagram of transfer learning study of ABIDE-I.}
	\label{transfer}
\end{figure}

\begin{figure*}
	\centering
	\includegraphics[scale=0.43]{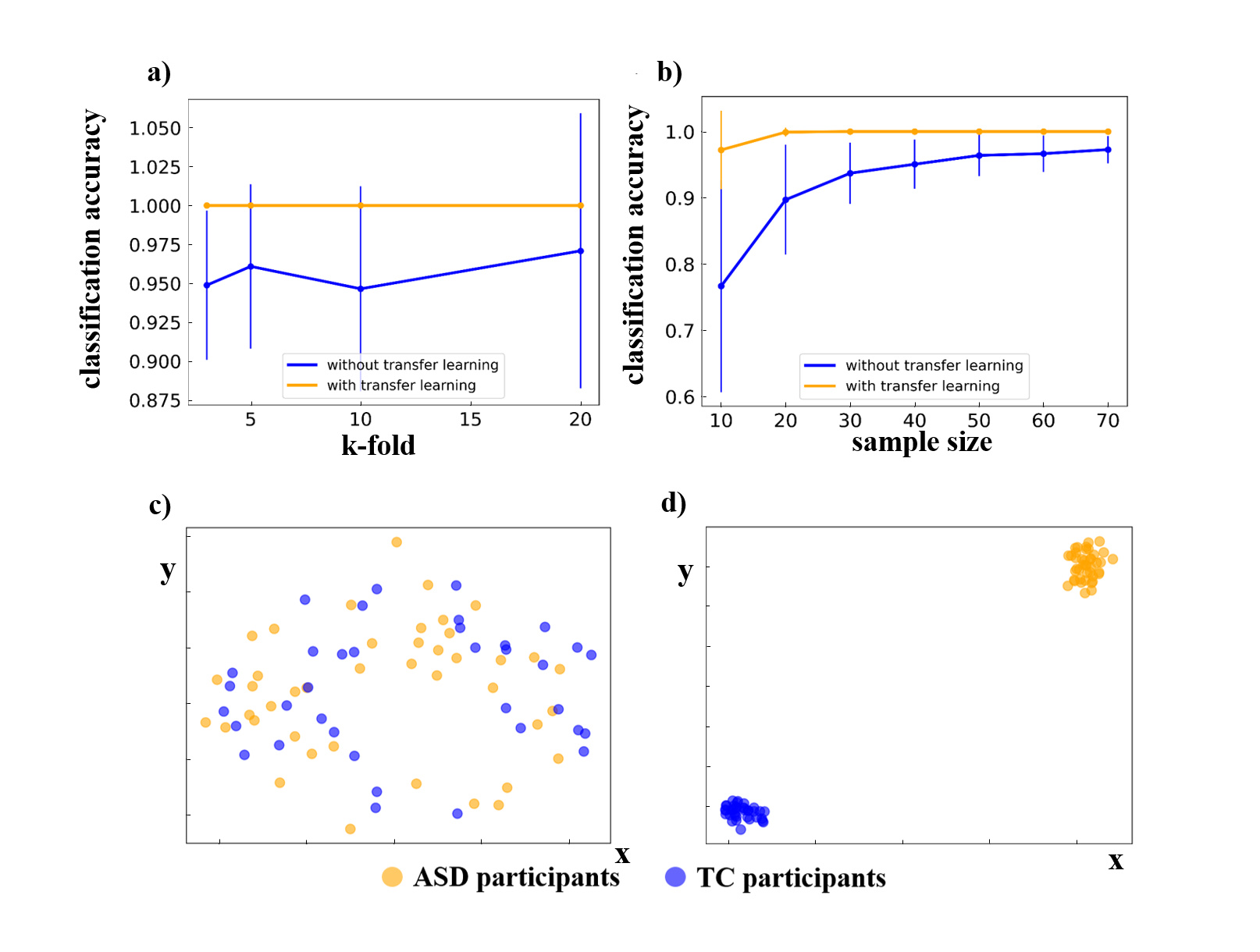}
	\caption{Classification results and the visualization  analysis of the features. a) accuracy of classifying ASD with/without transfer learning, where ASD participants are represented by ASD-specific features; b) relationship between classification accuracy and sample size with/without transfer learning, where ASD participants are represented by ASD-specific features; visualizing ASD participants and TC participants in 2-D feature space with TC participants being represented by common shared features while ASD participants being represented by either c) common shared features, or d) ASD-specific features.}
	\label{classification2}
\end{figure*}

As presented in Figure \ref{transfer}, we collect 982 s-MRIs from ABIDE-I, consisting of 470 ASD participants (age range 7-64 years) and 512 TC participants (age range 6-56 years), which are further used to train the contrastive variational autoencoder model, named as MABIDE. After the training, the ASD participants from Shenzhen Children's Hospital are passed through the MABIDE to acquire the ASD-specific features, and the TC participants from Shenzhen Children's Hospital are passed through the MABIDE to acquire common shared features. Only the ASD-specific features for ASD participants are explored here, as it's been demonstrated to be the predictive power for classifying ASD in the Section 3.1. The same random forest model is also trained to classify ASD for children in our work based on the features extracted from MABIDE, whose classification accuracy is tagged with "with transfer learning", as indicated in Figure \ref{classification2}.a and b. The K-fold number and sample size analysis are also conducted here, and the corresponding classification accuracies from Figure \ref{classification1}.a and b are recorded in Figure \ref{classification2}.a and b, tagged with "without transfer learning".  In Figure \ref{classification2}.a, the accuracy of ASD classification based on ASD-specific features is improved by transfer learning, regardless of K-fold numbers. Besides, as can be seen from Figure \ref{classification2}.b, the classification accuracy is less sensitive to data size after the transfer learning, and the model is highly predictive even with small training data. 

When visualizing ASD participants based on common shared features in Figure \ref{classification2}.c and ASD-specific features in Figure \ref{classification2}.d, with both TC participants as the background, we can see that the ASD participants and TC participants still overlap well with each other when both being represented using common shared features but the boundary between ASD participants and TC participants is clearer when ASD participants are represented by ASD-specific features (i.e., the ASD participants are perfectly separated from TC participants), explaining why the classification accuracy is perfectly 1.00 in Figure \ref{classification2}.a, regardless of K-fold numbers. From this study, ABIDE-I can be considered as a good source task for transfer learning, in terms of extracting effective features for classifying ASD in children. The other open-sourced databases may be explored and considered for similar purposes. 

The contrastive variational autoencoder is built to learn the capability of separating ASD-specific features from common-shared features in the brain. The developmental trajectory of the brain in individuals with ASD can explain that the structural differences between the ASD and TC brains are more pronounced in early-age children$^{[36]}$. The brain differences between ASD and TC individuals from ABIDE dataset are less apparent due to the older age of the participants, i.e., it's more difficult for contrastive variational autoencoder to separate ASD-specific features from common-shared features in adult's brain (the harder task). Thanks to the large amount of data contained in ABIDE, the model can do the harder task well. When the model is used in Shenzhen Children's Hospital dataset, which consists of younger individuals with more noticeable brain differences, it can separate ASD-specific features from common-shared features better, because this is an easier task.

\subsection{Neuroanatomical interpretation}
\noindent
Although the ASD-specific features have high predictive power in classifying ASD, they don't explicitly tell which regions they correspond to in the brain, causing no sense of interpretation. Knowing the specific brain regions associating with ASD is critical for ASD intervention in the future, as they serve as the potential biomarkers that can help target clinical treatment$^{[37]}$. To find the specific brain regions associating with ASD-specific features, we correlate them to the surface area that are manually extracted from the segmented cortex based on a specific atlas (more details can be checked in Section 2.2). The different cortical regions are responsible for different functions$^{[38]}$, and the malfunction of certain regions may lead to the behavior of ASD. The cortexes of children used in this study are first segmented into 34 regions, naming temporalpole, bankssts, rostralanteriorcingulate, supramarginal, inferiorparietal, posteriorcingulate, parsopercularis, lateralorbitofrontal, middletemporal, entorhinal, frontalpole, parstriangularis, paracentral, lateraloccipital, parahippocampal, inferiortemporal, pericalcarine, caudalmiddlefrontal, cuneus, lingual, fusiform, superiorfrontal, transversetemporal, superiortemporal, medialorbitofrontal, isthmuscingulate, precuneus, caudalanteriorcingulate, precentral, parsorbitalis, rostralmiddlefrontal, postcentral, insula, superiorparietal. We don't explore the functions of all the segmented regions here, rather focusing on the specific regions that distinguish ASD participants from TC participants most. 

We use representational similarity analysis (RSA)$^{[39]}$ to test whether the ASD-specific features/common shared features correlate with surface area of segmented cortical regions. We first calculate the pair-wise dissimilarity between participants with respect to the ASD-specific features and common shared features separately and obtain two groups of dissimilarity matrixes. Specifically, 42 s-MRIs of ASD participants from Shenzhen Children's Hospital are passed through two encoders of the trained CVAE models separately, then 42 ASD-specific features (42 $\times$ 16) and 42 common shared features (42 $\times$ 16) are generated by sampling from the encoded feature distributions. We calculate the pairwise Euclidean distances between ASD-specific features and between common shared features, which form two dissimilarity matrixes (42$\times$ 42) respectively. We sample 10 times from each encoded feature distribution and two groups of dissimilarity matrixes (10 for ASD-specific features and 10 for common shared features) are generated with method mentioned above. We then repeat this process for surface area of each segmented cortical region of ASD participants. Finally, we correlate the dissimilarity matrices of ASD-specific features/common shared features to the dissimilarity matrices of each cortical region's surface area using the Kendall rank correlation coefficient (Kendall $\tau$)$^{[40]}$, a measurement (value between -1 and 1) between two rankings $X$ ($x_{1}$, $x_{2}$,..., $x_{n}$) and $Y$ ($y_{1}$, $y_{2}$,..., $y_{n}$) using the formula: ($P$ - $Q$) / sqrt(($P$ + $Q$ + $T$) $\times$ ($P$ + $Q$ + $U$)), where $P$ is the number of concordant pairs (pairs that satisfy condition $x_{i}$ \textgreater $x_{j}$ \& $y_{i}$ \textgreater $y_{j}$ or pairs that satisfy condition $x_{i}$ \textless $x_{j}$ \& $y_{i}$ \textless $y_{j}$), $Q$ the number of discordant pairs (pairs that satisfy condition $x_{i}$ \textgreater $x_{j}$ \& $y_{i}$ \textless $y_{j}$ or pairs that satisfy condition $x_{i}$ \textless $x_{j}$ \& $y_{i}$ \textgreater $y_{j}$), $T$ the number of ties only in $X$, and $U$ the number of ties only in $Y$. If a tie occurs for the same pair in both $X$ and $Y$, it is not added to either $T$ or $U$. Values close to 1 indicate strong agreement of the correspondence between two rankings, while values close to -1 indicate strong disagreement. 

\begin{figure*}
	\centering
	\includegraphics[scale=0.235]{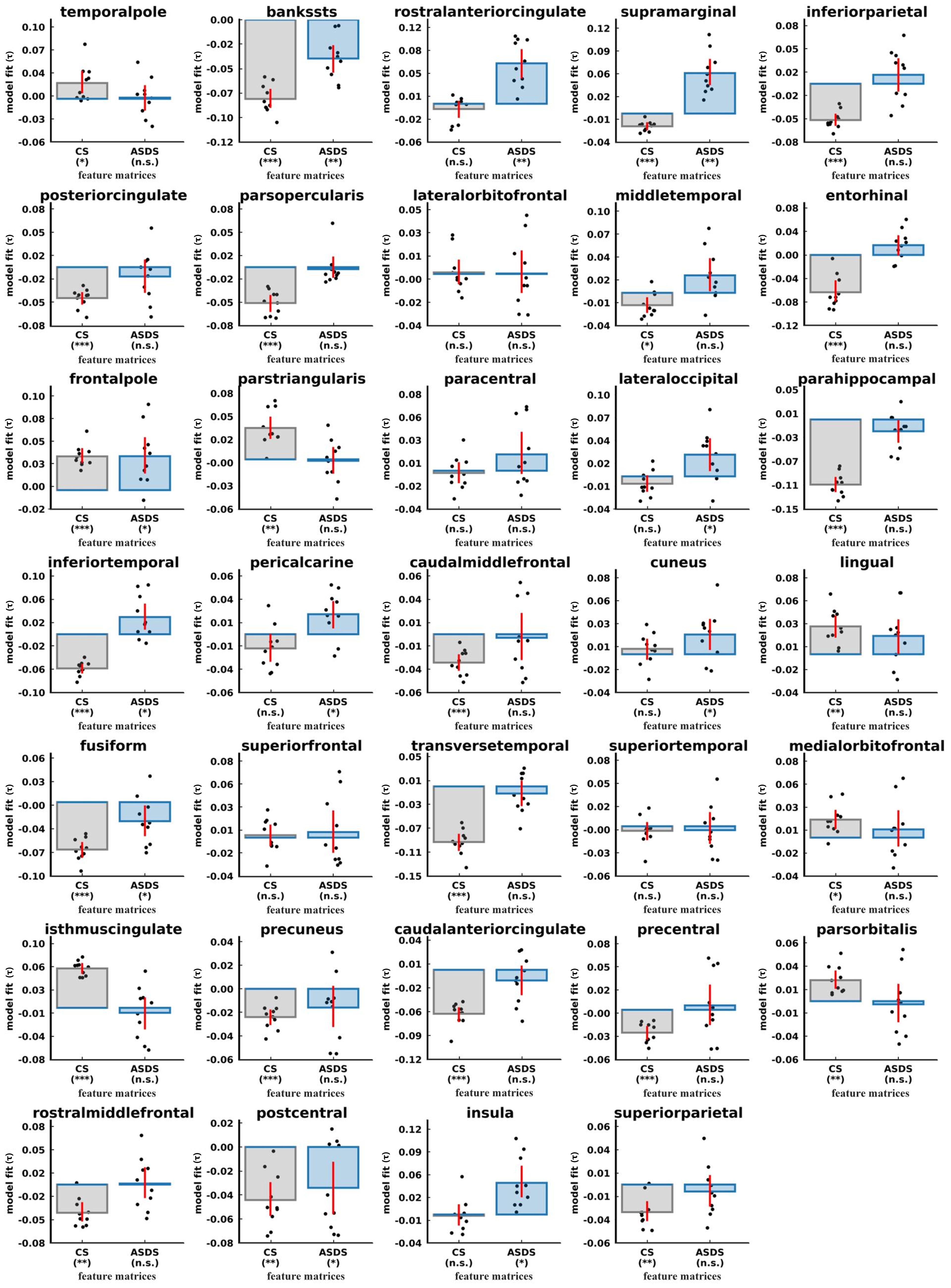}
	\caption{Kendall rank correlation coefficient (Kendall $\tau$) measuring agreement/disagreement between the dissimilarity matrices of ASD-specific (ASDS) features /common shared (CS) features and the matrices for each cortical region's surface area. The "n.s.", "*", "**", and "***" correspond to p-value smaller than 1, 5 $\times$ $10^{-2}$, $10^{-3}$, and $10^{-4}$.}
	\label{explain}
\end{figure*}

The significance of a relationship is characterized by p-value, with p-value smaller than 5 $\times$ $10^{-2}$ as significant. We pay attention to the cortical regions, where ASD-specific features have positive correlations and common shared features have negative correlations, both statistically significant. In other words, those cortical regions indicate the biggest brain differences between ASD participants and TC participants involved in our study. Based on Figure \ref{explain}, the names of cortical regions that can distinguish ASD participants from TC participants most, are \emph{supramarginal} and \emph{inferiortemporal}, and we name those regions ASD-specific-regions. The names and functions of ASD-specific-regions are recorded in Table \ref{table1}. Based on Table \ref{table1}, the malfunctions of ASD-specific-regions will make impact on your interpretation of spoken language as well as written language (phonological decisions and visual word recognition) and your recognition of objects in the environment (visual recognition of objects), which will further lead to social communication/interaction disorders. ASD is a complex group of developmental disorders characterized by difficulties in social communication/interaction that can persist throughout life$^{[44, 45]}$, and the ASD-specific-regions found above indicate where those problems come from in the brain, which serve as reasonable biomarkers that could help target treatment.

\begin{table*}
	\caption{Names and functions of the areas that distinguish ASD participants from TC participants most in our study.}
	\vskip 5mm
	\label{table1}
	\centering
		\begin{tabular}{lccccccccc}
			\toprule
			Name of area    &Function of area \\
			\midrule
			\emph{supramarginal}$^{[41, 42]}$   &phonological decisions and visual word recognition  \\
			\emph{inferiortemporal}$^{[43]}$  &visual recognition of objects \\
			
			\bottomrule
		\end{tabular}
\end{table*}

\section{CONCLUSION \& DISCUSSION}
\noindent
In summary, we conduct machine classification of ASD in children based on s-MRI features extracted using CVAE and achieve a stable predictive accuracy above 0.94 under different cross-validation scenarios. The predictive accuracy of the classification model is sensitive to data size, which will be degraded when lowering the data size. To mitigate the impact of data size on the predictive accuracy, we introduce a transfer learning strategy, which learns the knowledge from ABIDE I dataset and transfers the learned knowledge to improve the classification accuracy here when data is extremely limited (below 30). As demonstrated, the transfer learning works effectively towards making predictive accuracy less sensitive to data size. Finally, unlike most existing algorithms that only simply perform classification study of ASD, we also explain the classification results by analyzing the features using t-SNE visualization technique, what's more, we also conduct neuroanatomical interpretation via representational similarity analysis to disclose the specific cortical regions correlated to ASD, which may help find potential biomarkers for ASD treatments.

Although we have made some progress in ASD classification and neuroanatomical interpretation tasks, the lack of early-age ASD datasets is still hindering us to well validate the robustness of our algorithms, even though we have introduced ABIDE dataset to improve generalization performance as much as possible. In addition, the analysis of the learned feature space in this paper is not sufficiently in-depth. In future work, we will expand our early-age ASD dataset and explore the properties of the learned feature space more deeply by employing methods such as disentangling technique$^{[46, 47]}$, besides, we can also try to incorporate prior knowledge to enhance the capabilities of our framework.

\vskip 2mm
\large
\noindent
\textbf{Acknowledgment}
\vskip 2mm

\size{10pt}
\noindent
The authors acknowledge the financial support from Shenzhen Science and Technology Program under grant no. KQTD20200820113106007, the Strategic Priority Research Program of Chinese Academy of Sciences (grant no. XDB38050100), Shenzhen Key Laboratory of Intelligent Bioinformatics (ZDSYS20220422103800001), Shenzhen Basic Research Fund under grant no. RCYX20200714114734194, Key Research and Development Project of Guangdong Province under grant no. 2021B0101310002, National Natural Science Foundation of China under grant No. U22A2041, National Natural Science Foundation of China under grant no. 62272449, Shenzhen Science and Technology Program (Grant No. SGDX20201103095603009) and Youth Innovation Promotion Association (Y2021101), CAS.

\vskip 2mm
\normalsize
\noindent
\renewcommand\refname{\large

\begin{biography}[maruimin.jpg]
	\noindent
	\textbf{Ruimin Ma} Shenzhen Institute of Advanced Technology, Shenzhen, China. Ruimin Ma received his Ph.D. degree from the University of Notre Dame, USA in 2021. He is a postdoc research fellow at Shenzhen Institute of Advanced Technology, China. His main research interests are artificial intelligence, bioinformatics, etc..
\end{biography}

\begin{biography}[xieruitao.jpg]
	\noindent
	\textbf{Ruitao Xie} received the BS and MS degrees from Shenzhen University, Shenzhen, China in 2018 and 2021 respectively. He is currently working toward the PhD degree in the University of Chinese Academy of Sciences, Beijing, China. His research interests include deep learning, image processing and bioinformatics.
\end{biography}

\begin{biography}[wangyanlin.jpg]
	\noindent
	\textbf{Yanlin Wang} Shenzhen Institute of Advanced Technology, Shenzhen, China. Yanlin Wang received his Ph.D. degree from the University of Sichuan, China in 2021. He is a postdoc research fellow at Shenzhen Institute of Advanced Technology, China. His main research interests are cognitive and computational neuroscience, bioinformatics, etc..
\end{biography}

\begin{biography}[mengjintao.jpg]
	\noindent
	\textbf{Jintao Meng} is a associate researcher at Shenzhen Institutes of Advanced Technology, Chinese Academy of Sciences. He received the B.S. and M.S. degrees in computer science from the Central China Normal University, Wuhan, in 2005 and 2008 respectively and the Ph.D. degree in computer architecture from Institute of Computing Technology, Chinese Academy of Sciences, Beijing, in 2016.  His research interests include high performance computing, bioinformatics, and graph computing.
\end{biography}

\begin{biography}[weiyanjie.jpg]
	\noindent
	\textbf{Yanjie Wei} is a professor and the director in Center for High Performance Computing, Shenzhen Institute of Advanced Technology, Chinese Academy of Sciences. He earned his Ph.D in 2007 at Michigan Tech University in the field of computational biophysics. From 2008 to 2011, he worked as a postdoctoral research associate at Princeton University. His research focuses on high performance computing and computational biology/bioinformatics. 
\end{biography}
\begin{biography}[caiyunpeng.jpg]
	\noindent
	\textbf{Yunpeng Cai} received the PhD degree in computer science and technology from Tsinghua University, Beijing, China, in 2007. From 2007 to 2011, he was a postdoctoral researcher at University of Florida, USA. He is currently a professor at Shenzhen Institute of Advanced Technology, Chinese Academy of Sciences, Shenzhen, China. His research interests include health big data, health informatics, bioinformatics, and machine learning. 
\end{biography}

\begin{biography}[xiwenhui.jpg]
	\noindent
	\textbf{Wenhui Xi} received the BS and PhD degrees in applied physics from Nanjing University, China in 2007 and 2012, respectively. From 2012 to 2018, he was a postdoctoral researcher at Fudan University, China and University of Oklahoma, USA. He is currently an associate researcher at Shenzhen Institutes of Advanced Technology, Chinese Academy of Sciences. His research interests include biomedical big data bioinformatics and molecular simulation of proteins and peptides. 
\end{biography}

\begin{biography}[panyi.png]
	\noindent
	\textbf{Yi Pan} received the BEng (computer engineering) and MEng (computer engineering) degrees from Tsinghua University, Beijing, China in 1982 and 1984, respectively, and the PhD degree (computer science) from the University of Pittsburgh, USA in 1991. He is currently working as a dean and chair professor at the Faculty of Computer Science and Control Engineering, Shenzhen Institute of Advanced Technology, Chinese Academy of Sciences, Shenzhen, Guangdong, China. He has published more than 450 papers including over 250 journal papers with more than 100 papers published in IEEE/ACM Transactions/Journals. In addition, he has edited/authored 43 books. His work has been cited more than 24411 times based on Google Scholar and his current h-index is 95. He has served as an editor-in-chief or editorial board member for 20 journals including 7 IEEE Transactions. He is the recipient of many awards including one IEEE Transactions Best Paper Award, five IEEE and other international conference or journal Best Paper Awards, 4 IBM Faculty Awards, 2 JSPS Senior Invitation Fellowships, IEEE BIBE Outstanding Achievement Award, IEEE Outstanding Leadership Award, NSF Research Opportunity Award, and AFOSR Summer Faculty Research Fellowship. He has organized numerous international conferences and delivered keynote speeches at over 60 international conferences around the world. His research interests include parallel and distributed processing systems, Internet technology, and bioinformatics.
\end{biography}


\textbf{References}}

\begin{thebibliography}{99}
	\normalsize\addtolength{\itemsep}{-1em}
	\vspace {1.5mm}
	
	\bibitem[1]{1}
	Lord, C., Elsabbagh, M., Baird, G. and Veenstra-Vanderweele, J., Autism spectrum disorder. The lancet, 2018. 392(10146): p. 508-520.
	
	\bibitem[2]{2}
	Thabtah, F.a.P., D., A new machine learning model based on induction of rules for autism detection. Health informatics journal, 2020. 26(1): p. 264-286.
	
	\bibitem[3]{3}
	Moridian, P., Ghassemi, N., Jafari, M., Salloum-Asfar, S., Sadeghi, D., Khodatars, M., Shoeibi, A., Khosravi, A., Ling, S.H., Subasi, A. and Abdulla, S.A., Automatic autism spectrum disorder detection using artificial intelligence methods with MRI neuroimaging: A review. arXiv preprint arXiv, 2022. 2206(11233).
	
	\bibitem[4]{4}
	Di Martino, A., Yan, C.G., Li, Q., Denio, E., Castellanos, F.X., Alaerts, K., Anderson, J.S., Assaf, M., Bookheimer, S.Y., Dapretto, M. and Deen, B., The autism brain imaging data exchange: towards a large-scale evaluation of the intrinsic brain architecture in autism. Molecular psychiatry, 2014. 19(6): p. 659-667.
	
	\bibitem[5]{5}
	Di Martino, A., O'connor, D., Chen, B., Alaerts, K., Anderson, J.S., Assaf, M., Balsters, J.H., Baxter, L., Beggiato, A., Bernaerts, S. and Blanken, L.M., Enhancing studies of the connectome in autism using the autism brain imaging data exchange II. Scientific data, 2017. 4(1): p. 1-15.
	
	\bibitem[6]{6}
	Kong, Y., Gao, J., Xu, Y., Pan, Y., Wang, J. and Liu, J., Classification of autism spectrum disorder by combining brain connectivity and deep neural network classifier. Neurocomputing, 2019. 324: p. 63-68.
	
	\bibitem[7]{7}
	Hiremath, Y., Ismail, M., Verma, R., Antunes, J. and Tiwari, P., Combining deep and hand-crafted MRI features for identifying sex-specific differences in autism spectrum disorder versus controls. In Medical Imaging 2020: Computer-Aided Diagnosis, 2020, March. 11314: p. 445-451.
	
	\bibitem[8]{8}
	Rohlfing, T., Zahr, N.M., Sullivan, E.V. and Pfefferbaum, A., The SRI24 multichannel atlas of normal adult human brain structure. Human brain mapping, 2010. 31(5): p. 798-819.
	
	\bibitem[9]{9}
	Gao, J., Chen, M., Li, Y., Gao, Y., Li, Y., Cai, S. and Wang, J., Multisite autism spectrum disorder classification using convolutional neural network classifier and individual morphological brain networks. Frontiers in Neuroscience, 2021. 14.
	
	\bibitem[10]{10}
	Yang, R., Ke, F., Liu, H., Zhou, M. and Cao, H.M., Exploring sMRI biomarkers for diagnosis of autism spectrum disorders based on multi class activation mapping models. IEEE Access, 2021. 9(124122-124131).
	
	\bibitem[11]{11}
	Zhou, B., Khosla, A., Lapedriza, A., Oliva, A. and Torralba, A., Learning deep features for discriminative localization. In Proceedings of the IEEE conference on computer vision and pattern recognition, 2016: p. 2921-2929.
	
	\bibitem[12]{12}
	Selvaraju, R.R., Cogswell, M., Das, A., Vedantam, R., Parikh, D. and Batra, D., Grad-cam: Visual explanations from deep networks via gradient-based localization. In Proceedings of the IEEE international conference on computer vision, 2017: p. 618-626.
	
	
	\bibitem[13]{13}
	Mishra, M.a.P., U.C., A classification framework for Autism Spectrum Disorder detection using sMRI: Optimizer based ensemble of deep convolution neural network with on-the-fly data augmentation. Biomedical Signal Processing and Control, 2023. 84.
	
	\bibitem[14]{14}
	Sharif, H.a.K., R.A., A novel machine learning based framework for detection of autism spectrum disorder (ASD). Applied Artificial Intelligence, 2022. 36(1): p. 2004655.
	
	\bibitem[15]{15}
	Wang, Z., Peng, D., Shang, Y. and Gao, J., Autistic spectrum disorder detection and structural biomarker identification using self-attention model and individual-level morphological covariance brain networks. Frontiers in Neuroscience, 2021.
	
	\bibitem[16]{16}
	Ma, R., Wang, Y., Wei, Y. and Pan, Y., Meta-data Study in Autism Spectrum Disorder Classification Based on Structural MRI. arXiv preprint arXiv, 2022. 2206(05052).
	
	\bibitem[17]{17}
	Sarovic, D., Hadjikhani, N., Schneiderman, J., Lundström, S. and Gillberg, C., Autism classified by magnetic resonance imaging: A pilot study of a potential diagnostic tool. International journal of methods in psychiatric research, 2020. 29(4): p. 1-18.
	
	\bibitem[18]{18}
	Conti, E., Retico, A., Palumbo, L., Spera, G., Bosco, P., Biagi, L., Fiori, S., Tosetti, M., Cipriani, P., Cioni, G. and Muratori, F., Autism Spectrum Disorder and Childhood Apraxia of Speech: Early language-related hallmarks across structural MRI study. Journal of personalized medicine, 2020. 10(4).
	
	\bibitem[19]{19}
	Zhang, X., Ding, X., Wu, Z., Xia, J., Ni, H., Xu, X., Liao, L., Wang, L. and Li, G., Siamese verification framework for autism identification during infancy using cortical path signature features. In 2020 IEEE 17th International Symposium on Biomedical Imaging (ISBI), 2020, April: p. 1-4.
	
	\bibitem[20]{20}
	National Database for Autism Research (NDAR). https://ndar.nih.gov.
	
	\bibitem[21]{21}
	Zöllei, L., Iglesias, J.E., Ou, Y., Grant, P.E. and Fischl, B., Infant FreeSurfer: An automated segmentation and surface extraction pipeline for T1-weighted neuroimaging data of infants 0-2 years. Neuroimage, 2020. 218.
	
	\bibitem[22]{22}
	Avants, B.B., Duda, J.T., Kilroy, E., Krasileva, K., Jann, K., Kandel, B.T., Tustison, N.J., Yan, L., Jog, M., Smith, R. and Wang, Y., Large-scale evaluation of ANTs and FreeSurfer cortical thickness measurements. Scientific Data, 2014. 2.
	
	\bibitem[23]{23}
	FreeSurfer, F.B., FreeSurfer. NeuroImage, 2012. 62(2): p. 774-781.
	
	\bibitem[24]{24}
	Desikan, R.S., Ségonne, F., Fischl, B., Quinn, B.T., Dickerson, B.C., Blacker, D., Buckner, R.L., Dale, A.M., Maguire, R.P., Hyman, B.T. and Albert, M.S., An automated labeling system for subdividing the human cerebral cortex on MRI scans into gyral based regions of interest. Neuroimage, 2006. 31(3): p. 968-980.
	
	\bibitem[25]{25}
	Severson, K.A., Ghosh, S. and Ng, K., Unsupervised learning with contrastive latent variable models. In Proceedings of the AAAI Conference on Artificial Intelligence, 2019, July. 33(01): p. 4862-4869.
	
	\bibitem[26]{26}
	Abid, A.a.Z., J., 2019. Contrastive variational autoencoder enhances salient features. arXiv preprint arXiv:1902.04601., Contrastive variational autoencoder enhances salient features. arXiv preprint arXiv, 2019. 1902(04601).
	
	\bibitem[27]{27}
	Kingma D P, Ba J. Adam: A method for stochastic optimization[J]. arXiv preprint arXiv:1412.6980, 2014.
	
	\bibitem[28]{28}
	Breiman, L., Random forests. Machine Learning, 2001. 45(1): p. 5-32.
	
	\bibitem[29]{29}
	Van der Maaten, L.a.H., G., Visualizing data using t-SNE. Journal of machine learning research, 2008. 9(11).
	
	\bibitem[30]{30}
	Jönemo J, Abramian D, Eklund A. Evaluation of augmentation methods in classifying autism spectrum disorders from fMRI data with 3D convolutional neural networks[J]. Diagnostics, 2023, 13(17): 2773.
	
	
	\bibitem[31]{31}
	Nogay H S, Adeli H. Diagnostic of autism spectrum disorder based on structural brain MRI images using, grid search optimization, and convolutional neural networks[J]. Biomedical Signal Processing and Control, 2023, 79: 104234.
	
	
	
	
	\bibitem[32]{32}
	Aglinskas, A., Hartshorne, J.K. and Anzellotti, S., Contrastive machine learning reveals the structure of neuroanatomical variation within autism. Science, 2022. 376(6597): p. 1070-1074.
	
	\bibitem[33]{33}
	Khadem-Reza, Z.K. and H. Zare, Evaluation of brain structure abnormalities in children with autism spectrum disorder (ASD) using structural magnetic resonance imaging. The Egyptian Journal of Neurology, Psychiatry and Neurosurgery, 2022. 58(1).
	
	\bibitem[34]{34}
	Weiss, K., Khoshgoftaar, T.M. and Wang, D., A survey of transfer learning. Journal of Big data, 2016. 3(1): p. 1-40.
	
	\bibitem[35]{35}
	Ma, R., Y.J. Colon, and T. Luo, Transfer Learning Study of Gas Adsorption in Metal-Organic Frameworks. ACS Appl Mater Interfaces, 2020. 12(30): p. 34041-34048.
	
	\bibitem[36]{36}
	Courchesne E, Pramparo T, Gazestani V H, et al. The ASD Living Biology: from cell proliferation to clinical phenotype[J]. Molecular psychiatry, 2019, 24(1): 88-107.
	
	\bibitem[37]{37}
	Goldani, A.A., Downs, S.R., Widjaja, F., Lawton, B. and Hendren, R.L., Biomarkers in autism. Frontiers in psychiatry, 2014. 5.
	
	\bibitem[38]{38}
	Mercadante, A.A.a.T., P., Neuroanatomy, Gray Matter. 2020.
	
	\bibitem[39]{39}
	Kriegeskorte, N., M. Mur, and P. Bandettini, Representational similarity analysis - connecting the branches of systems neuroscience. Front Syst Neurosci, 2008. 2: p. 4.
	
	\bibitem[40]{40}
	Abdi, H., The Kendall rank correlation coefficient. Encyclopedia of Measurement and Statistics, 2007. Sage, Thousand Oaks, CA: p. 508-510.
	
	
	\bibitem[41]{41}
	Hartwigsen, G., Baumgaertner, A., Price, C.J., Koehnke, M., Ulmer, S. and Siebner, H.R., Phonological decisions require both the left and right supramarginal gyri. Proceedings of the National Academy of Sciences, 2010. 107(38): p. 16494-16499.
	
	\bibitem[42]{42}
	Stoeckel, C., Gough, P.M., Watkins, K.E. and Devlin, J.T., Supramarginal gyrus involvement in visual word recognition. Cortex, 2009. 45(9): p. 1091-1096.
	
	\bibitem[43]{43}
	Miyashita, Y., Inferior temporal cortex: where visual perception meets memory. Annual review of neuroscience, 1993. 16(1): p. 245-263.
	
	\bibitem[44]{44}
	Association, D.-A.P., Diagnostic and statistical manual of mental disorders. Arlington: American Psychiatric Publishing, 2013. 10.
	
	\bibitem[45]{45}
	Hyman, S.L., Levy, S.E., Myers, S.M., Kuo, D.Z., Apkon, S., Davidson, L.F., Ellerbeck, K.A., Foster, J.E., Noritz, G.H., Leppert, M.O.C. and Saunders, B.S., Identification, evaluation, and management of children with autism spectrum disorder. Pediatrics, 2020. 145(1).
	
	\bibitem[46]{46}
	Lang O, Gandelsman Y, Yarom M, et al. Training a GAN to explain a classifier in StyleSpace[C]//Proc. IEEE/CVF Conference on Computer Vision and Pattern Recognition (CVPR)(ed O'Conner, L.). 2021: 673-682.
	
	\bibitem[47]{47}
	Atad M, Dmytrenko V, Li Y, et al. Chexplaining in style: Counterfactual explanations for chest x-rays using stylegan[J]. arXiv preprint arXiv:2207.07553, 2022.
	
\end{thebibliography}
\end{document}